# Recognizing Hand Use and Hand Role at Home After Stroke from Egocentric Video

Meng-Fen Tsai[1,3,6], Rosalie H. Wang[2,3,5,6], and José Zariffa[1,3,4,5,6]

[1] Institute of Biomedical Engineering, University of Toronto
[2] Department of Occupational Science and Occupational Therapy, University of Toronto
[3] KITE, Toronto Rehabilitation Institute, University Health Network
[4] Edward S. Rogers Sr. Department of Electrical and Computer Engineering, University of Toronto
[5] Rehabilitation Sciences Institute, University of Toronto
[6] Robotics Institute, University of Toronto

**Abstract**
**Introduction**: Hand function is a central determinant of independence after stroke. Measuring hand use in the home environment is necessary to evaluate the impact of new interventions, and calls for novel wearable technologies. Egocentric video can capture hand-object interactions in context, as well as show how more-affected hands are used during bilateral tasks (for stabilization or manipulation). Automated methods are required to extract this information. **Objective**: To use artificial intelligence-based computer vision to classify hand use and hand role from egocentric videos recorded at home after stroke. **Methods**: Twenty-one stroke survivors participated in the study. A random forest classifier, a SlowFast neural network, and the Hand Object Detector neural network were applied to identify hand use and hand role at home. Leave-One-Subject-Out-Cross-Validation (LOSOCV) was used to evaluate the performance of the three models. Between-group differences of the models were calculated based on the Mathews correlation coefficient (MCC). **Results**: For hand use detection, the Hand Object Detector had significantly higher performance than the other models. The macro average MCCs using this model in the LOSOCV were 0.50 ± 0.23 for the more-affected hands and 0.58 ± 0.18 for the less-affected hands. Hand role classification had macro average MCCs in the LOSOCV that were close to zero for all models. Conclusion: Using egocentric video to capture the hand use of stroke survivors at home is feasible. Pose estimation to track finger movements may be beneficial to classifying hand roles in the future.

## I. Introduction

Upper limb function is a determinant of quality of life after stroke [1]. More than 65% of stroke survivors have remaining upper limb impairments six months after stroke [2, 3]. Hemiplegia or hemiparesis is a common motor deficit after stroke that causes the more-affected limbs to experience difficulties isolating or executing movements. Novel interventions for upper limb function are required to improve independence in activities of daily living (ADLs). Prior to translating a new intervention into practice, its ultimate impact on the daily life of stroke survivors should be quantified through appropriate outcome measures. The upper limb function measured in a clinical setting is not always demonstrated in daily life [4-9]. According to the International Classification of Functioning, Disability and Health (ICF) from the World Health Organization, measuring function in a hospital and in the community corresponds to two different domains of function - capacity and performance, respectively [10]. The former measures an individual's highest level of function and the latter captures the functional performance in their usual environment.

Most studies measuring upper limb performance have focused on reporting arm use rather than hand use. Reaching and grasping are distinct components required to complete a task. Manipulating an object can be carried out in various ways, depending on the level of impairments in hand function. Investigating hand use in addition to arm use can reveal different aspects of upper limb function after stroke. In clinical or laboratory environments, arm and hand function are assessed in different subsets of upper limb assessments, such as the Action Research Arm Test (ARAT) [11] and the Fugl-Meyer assessment for upper extremity (FMA-UE) [12]. Therefore, there is a need for assessments that focus on the performance aspect of hand function in home and community environments.

In addition to hand use, the role of a more-affected hand during a bimanual ADL is another important indicator of hand function for individuals with hemiplegia. Stabilizer and manipulator are the two hand roles defined in the Chedoke Arm and Hand Activity Inventory [13]. The role of the more-affected hand depends on the severity of hand impairment and whether the affected hand was dominant pre-stroke. Most studies relevant to distinguishing manipulation and stabilization explored type of hand grasp

[14], hand posture tracking [15, 16], and upper limb assessment score estimation [17] rather than identifying hand roles. To date, automated methods to identify the hand roles of stroke survivors during bimanual ADLs have not yet been explored.

Wearable devices have been proposed to capture upper limb function in the community, such as accelerometers [18-21], magnetometers (Manumeter) [22, 23], force myography (TENZR) [24], and wearable cameras [25-27]. Wrist-worn accelerometers capture arm use rather than hand use [19, 28]. The TENZR is a wristband with a force myography sensor that measures surface force to capture reach-to-grasp movements [24]. However, the TENZR was reported to potentially not be able to detect small movements and cannot distinguish grasp types [29]. Finger-worn accelerometers [17, 30, 31] and the Manumeter [32, 33] record hand use, however, devices worn on a finger may affect naturalistic hand movements during ADLs. In addition, the recordings of the Manumeter may be impacted by metal objects, such as doorknobs and utensils, which are common household objects [32, 33]. A first-person camera (egocentric camera) can capture hand use without interfering with hand movements. In addition, this modality provides context about the ADLs taking place and can reveal more clinical information such as hand postures. Although some large movements may be out of frame, most daily tasks are carried out between the waist and shoulder and can be recorded clearly [34]. However, the high complexity of egocentric video data makes automated analysis a challenge.

Machine learning-based computer vision approaches have previously been applied to capture context and hand movements from egocentric videos, such as action recognition [35, 36], human-robot interaction [37, 38], and pose estimation [39-41]. Some obstacles to extracting hand movements using the videos have been identified, such as unstable light sources and camera motion [42], blurriness due to fast hand movements, and hands being occluded by users or manipulated objects [43]. These obstacles may impede the successful classification of hand use and hand role after stroke. Hand-object interactions comprise the majority of hand use scenarios, and computer vision has previously been applied to capture these interactions. A hand contact was referred to as a hand-object interaction in some studies. Studies investigating the interactions revealed that key features of hand use included spatial and temporal information [44-46] and the common region involving a hand and a manipulated object [25, 47]. Hand region alone was reported insufficient to detect a hand contact [48, 49]. The application of these approaches to solve rehabilitation problems has not been explored. Therefore, the objective of this study was to use egocentric video to automatically capture hand use and hand roles at home after stroke. While our previous work addressed these problems in video recorded in a laboratory setting, to our best knowledge, this is the first study that sought to capture hand use and hand role of stroke survivors in real home environments using egocentric video.

## II. METHODS

### A. Participants

Stroke survivors were invited to participate in the study, which was approved by the Research Ethics Board of the University Health Network. Informed consent from participants and their caregivers (if involved) were obtained before enrollment into the study. The inclusion criteria for study participants were the following: 1) at least six months post-stroke; 2) self-reported difficulty in daily life due to an impairment of the more-affected hand; 3) impaired but not absent hand function, defined as a total ARAT score above 10 [50]; 4) Montreal Cognitive Assessment (MoCA) above 21, to avoid potential cognitive difficulties [51]; 5) no subluxation or significant pain when using their upper limbs; 6) no other neuromusculoskeletal disease affecting upper limb movements other than stroke.

### B. Study Protocol

Each participant had two visits to a home simulation laboratory at the KITE Research Institute. In the first visit, informed consent was obtained and clinical assessments, including the FMA-UE, the ARAT, the MoCA, and the Motor Activity Log-30, were carried out to ensure participant eligibility and provide an overview of their hand function. In the second visit, the researcher demonstrated the procedure for using a head-mounted egocentric camera (GoPro Hero 5, GoPro Inc., CA, USA), following a previously reported protocol [34]. After the demonstration, participants familiarized themselves with using the camera. They subsequently carried out a list of daily tasks (Appendix) in six different room settings in the laboratory (i.e., living room, dining room, bedroom, washroom, kitchen, and hallway), while recording egocentric videos. Participants were asked to carry out the tasks as they normally would (i.e., the grasp type was not constrained). The researcher discussed with the participants their daily routines to agree on representative ADLs for home recordings. Participants were encouraged to record on different days or at different times during a day to capture diverse activities. After the two study visits, participants self-recorded their daily routines at home to collect three sessions of 1.5 hour-long recordings and returned the videos to the researcher.

### C. Datasets

Two types of videos were collected in this study: egocentric videos recorded in the home simulation laboratory and at home. Selected tasks from the laboratory-based videos are referred to as the HomeLab dataset and selected tasks from the home recordings are referred to as the Home dataset. The tasks in the datasets were selected to incorporate diversity in hand-object interactions, such as carrying out different tasks or a similar task in different ways. In addition to the varied tasks and grasps, in both datasets, two non-interaction tasks (negative instances) per participant were selected in which hands were present but not manipulating any object, in order to balance the datasets. An instance includes only one hand and is treated as an independent training sample. One frame may contain two or more instances depending on the number of hands. The HomeLab dataset is used in this study as a supplementary set to investigate whether including tasks with a set of objects in a standardized environment can help to identify the hand-object interactions and hand roles in uncontrolled home environments. All the videos were recorded at 1280x720 resolution with 30 frames per second and analyzed at 720x405 resolution.

In the Home dataset, each participant had three or four tasks self-recorded at home and one of the tasks contained only negative instances (no interactions), such as a hand resting or swinging during walking, to balance the datasets. The HomeLab dataset included nine participants, P01-09, each one having at least eight tasks performed in the six room settings in the laboratory and of which two were negative tasks where a hand was waved quickly in the air. The hand-object interactions in every task and hand roles in bimanual tasks were manually annotated frame-by-frame in the datasets. The definition of a hand-object interaction is the manipulation of an object by the hand(s) for a functional purpose. As for the two hand roles in bimanual tasks, a stabilizer is defined as the hand being statically in contact with an object without changing the contact area between them (a static interaction) and a manipulator is defined as the hand moving an object with the contact area changing over time (a dynamic interaction). The inter-rater reliabilities of the hand-object interaction and hand role labels were investigated using Cohen's Kappa statistic [52]. The prevalence-adjusted bias-adjusted kappa (PABAK) was also reported in order to consider the prevalence of the hand-object interactions and hand roles in the datasets and the bias between the annotators. A Kappa coefficient between 0.61-0.8 is substantial and between 0.81-1 is almost perfect agreement [53].

### D. Hand-Object Interaction Detection and Hand Role Classification

Binary hand-object interaction detection and hand role classification were investigated using three models: a random forest classifier using manually selected features [25], a SlowFast Network [54], and the Hand Object Detector [55]. All the tasks in the datasets were used for the interaction detection, and only bimanual tasks were included for the hand role classification. In order to compare the performances between the models, all the hand bounding boxes for each model were the same and generated from the Hand-Object Detector. Hands from caregivers might be recorded and recognized as users' hands, which is a weakness of the model that cannot differentiate one from the other. If a hand had two predictions (e.g. two right hands in the frame), the final prediction was calculated by averaging the predictions for that hand. An average $\geq 0.5$ was considered as an interaction or manipulation. The implementation details of each model are in the following paragraphs.

### 1. Random Forest Classifier

This model used a binary random forest classifier with 150 trees for the interaction and hand role classifications, as previously reported in [25], where it was applied to a subset of the HomeLab data. Here, the hand detection and data post-processing steps in that study were removed and the former was replaced by using the hand bounding boxes generated from the Hand Object Detector. The pipeline for using the random forest classifier included hand segmentation, feature extraction, and the binary classifications of the interaction and hand roles (Figure 1). The hand segmentation differentiated the regions of hand and non-hand areas in a hand bounding box using UNET [56]. Colour, motion, and hand shape features were applied to the two classifications. One additional feature that reflected manipulation movements, the pixel-wise changes in hand size in ten subsequent frames, was used for the hand role classification. The colour features were the differences in the Hue, Saturation and Value (HSV) colour space histograms between the hand region, the region surrounding the hand (non-hand region), and the background. The motion features used the differences in magnitudes and directions in optical flow histograms between these regions. The hand shape feature was a Histogram of Oriented Gradients (HOG) within the bounding box. The details regarding the features can be found in [25]. A hand that had no predicted bounding box was categorized as no interaction and no hand role (neither a manipulation nor a stabilization).

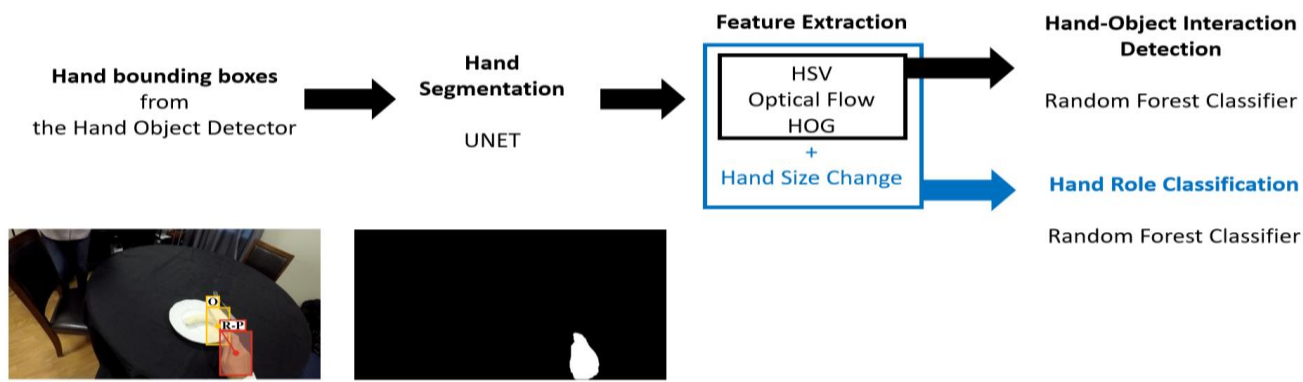

Fig. 1. Pipeline for using a binary random forest classifier with manually selected features to identify the hand-object interactions and the hand roles of stroke survivors. Colour (HSV), motion (optical flow), and hand shape (HOG) were common features of the interaction and hand role classifiers. Hand size change was an additional feature for the hand roles. The hand bounding box (red) was generated from the Hand Object Detector and R-P here represented that the right hand was in contact with a portable object. (HSV: Hue, Saturation, and Value; HOG: Histogram of Oriented Gradients).

### 2. SlowFast Network

SlowFast, a two-stream 3D convolutional neural network architecture, was used for the second model. The network was retrained from an activity recognition model trained on the MECCANO dataset [46]. The frame rate reduction ratio (alpha) and channel reduction ratio (beta) of the network were set as 4 and 8 between the slow and fast pathways [54]. A hand instance was defined as a set of bounding box regions with a size of 256x256 pixels over 32 frames for the interaction detection and 16 frames for the hand role classification. The base learning rate and weight decay were set at 0.0001 and 0.00001, which were the same as in [46]. Stochastic gradient descent (SGD) and cross entropy were used as the optimizer and loss function during training and the batch size was 7. The validation was carried out every 5 epochs to evaluate the network performance. The training was stopped when the training losses converged and validation error rate stopped decreasing over 10 epochs. The trained network with the minimum validation loss was applied to the testing set.

In order to compare the performance with frame-wise predictions from the random forest classifier, the prediction for a given 32- or 16-frame instance was applied as the predictions for all frames in the instance. Consecutive instances had 50% overlap, providing two predictions for each hand. If either prediction was an interaction, the final classification was as an interaction. The statistical results were calculated using the final prediction for each hand.

### 3. Hand Object Detector

The Hand Object Detector is a Faster Region-based Convolutional Neural Network (RCNN) algorithm that detects hand contacts, including self-contact, no contact, other person contact, portable object contact, and non-portable object contact [55]. In addition to predicting the contacts, object detection algorithms are included in the model, which generates hand and manipulated object bounding boxes in each frame. For the interaction detection, the Hand Object Detector was applied as baseline results without retraining. Portable object contact predictions were categorized as hand-object interactions, otherwise, as no interaction. For the hand role classification, the features of the last fully connected layer in the model were fed into another fully connected layer to classify the stabilizer and manipulator roles. A validation was carried out every 5 epochs. SDG and cross entropy were used as the optimizer and loss function during the training of the hand role classification, as for the SlowFast network, and the batch size was 20. The base learning rate and momentum were set as 0.0001 and 0.9. A learning rate schedule with weight decay of 0.001 was triggered if the training loss stopped decreasing for 2 epochs. The training stopped when the validation loss converged and stopped decreasing over 10 epochs. The trained hand role classification model with the minimum validation loss was applied to the testing set.

### E. Leave-One-Subject-Out-Cross-Validation (LOSOCV)

A leave-one-subject-out-cross-validation (LOSOCV) [57] was carried out to evaluate the performance for interaction detection and hand role classification in an unseen subject using each model in two conditions: using the Home dataset only and using both datasets. The two conditions aimed to investigate whether including a dataset of laboratory-based interactions (the HomeLab dataset) could improve the performance of the interaction detection and the hand role classification. The LOSOCV had the included dataset(s) split into testing, validation, and training sets. The testing set contained all the tasks performed by one single participant from the Home dataset and a validation set included one bimanual home task from each participant except for the one in the testing set. The rest of the tasks were in the training set. Neither the training nor validation sets included task performed by the tested participant, which was an unseen subject here. Macro and micro averages [58] of Matthews correlation coefficients (MCCs), F1-scores, precisions, recalls, and accuracies in each condition were reported to give a holistic evaluation of each model's performance. The macro average results provide the average performance of a model across individual tested participants and the micro average results reveal the overall model performance on all the participants. A repeated measures analysis of variance (ANOVA) or a Friedman test, depending on the data distribution, was applied to evaluate between-group differences of the MCC and the F1-score of each participant between the three models in the condition with highest macro average MCCs.

TABLE I
DEMOGRAPHIC INFORMATION AND TOTAL FUGL-MEYER ASSESSMENT FOR UPPER EXTREMITY (FMA-UE) SCORES OF PARTICIPANTS (N=21).

| Participant ID | Age (years) | Sex | Time after Onset of Stroke | Total Score of FMA-UE* (out of 66) |
|---|---|---|---|---|
| P01 | 83 | Male | 34 | 27 |
| P02 | 68 | Male | 4 | 27 |
| P03 | 66 | Male | 1 | 56 |
| P04 | 33 | Female | 18 | 24 |
| P05 | 48 | Female | 7 | 47 |
| P06 | 41 | Male | 1 | 66 |
| P07 | 74 | Male | 3 | 60 |
| P08 | 63 | Female | 2 | 37 |
| P09 | 64 | Male | 0.83 (10 months) | 66 |
| P14 | 60 | Male | 2 | 58 |
| P15 | 44 | Male | 2 | 66 |
| P16 | 60 | Male | 1 | 52 |
| P17 | 69 | Female | 1 | 66 |
| P18 | 35 | Female | 2 | 66 |
| P19 | 64 | Male | 2 | 66 |
| P20 | 59 | Female | 3 | 65 |
| P21 | 69 | Male | 2 | 60 |
| P22 | 50 | Male | 2 | 55 |
| P23 | 69 | Male | 1 | 61 |
| P25 | 76 | Male | 1 | 39 |
| P26 | 70 | Male | 1 | 64 |
| Mean ± SD | 60.2 ± 13.6 | - | 4.3 ± 7.8 | 53.7 ± 14.4 |

* FMA-UE score <25: severe, 26-50: moderate, and >50: mild upper limb impairment [59].

## III. RESULTS

Twenty-one stroke survivors, 15 males and 6 females, completed the study. The upper limb impairment levels of participants spanned across mild, moderate, and severe, according to the total FMA-UE score defined in [59]. The demographic information is provided in Table I. Some examples of Home and HomeLab instances are shown in Figures 2 and 3.

For the hand-object interaction detection, the Home and HomeLab datasets had a total of 79,543 frames (63% interaction and 37% no interaction) and 51,935 frames (53% interaction and 47% no interaction) labeled, respectively. The hand role classification only utilized the frames labeled as interactions in bimanual tasks since the roles were more representative of hand function level when two hands were involved. The labeled Home and HomeLab datasets for the hand role classification consisted of 64,291 frames (20% manipulation and 80% stabilization) and 32,480 frames (50% manipulation and 50% stabilization), respectively. The average numbers of instances in the testing, validation, and training sets in the LOSOCV conditions are listed in the results for each model.

### A. Inter-Rater Reliability of the Hand-Object Interaction and the Hand Roles

Five annotators manually labeled hand-object interactions and two of them labeled hand roles. Annotator 1 was the instructor for annotation and the inter-rater reliability was calculated between this annotator and the others. For the hand-object interactions, the PABAKs between annotators were all above 0.75 (Table II) and demonstrated that the inter-rater reliability for the hand-object interactions was at least substantial agreement. For the hand roles, the inter-rater reliability between Annotator 1 and 5 was 0.85 and demonstrated almost perfect agreement.

TABLE II
INTER-RATER RELIABILITY OF THE HAND-OBJECT INTERACTION AND HAND ROLE ANNOTATIONS.

| Annotation | Annotator | Number of Observations | Kappa | PABAK* |
|---|---|---|---|---|
| Hand-Object Interaction | 1 and 2 | 46,164 | 0.84 | 0.84 |
|  | 1 and 3 | 34,784 | 0.78 | 0.80 |
|  | 1 and 4 | 1,698 | 0.75 | 0.76 |
|  | 1 and 5 | 1,318 | 0.92 | 0.92 |
| Hand Role | 1 and 5 | 1,062 | 0.77 | 0.85 |

*PABAK: prevalence-adjusted bias-adjusted kappa (0.61-0.8: substantial agreement; > 0.81: almost perfect agreement).

### B. Hand-Object Interaction Detection

#### 1. Random Forest Classifier

The testing sets in the two conditions were the same and the average number of instances was $5,453 \pm 4,375$ instances ($75\% \pm 13\%$ interaction and $25\% \pm 13\%$ no interaction). The average numbers of training instances using the Home and both datasets were $109,063 \pm 4,375$ instances ($76\% \pm 1\%$ interaction and $24\% \pm 1\%$ no interaction) and $174,917 \pm 6,550$ instances ($71\% \pm 1\%$ interaction and $29\% \pm 1\%$ no interaction), respectively. A validation set was not applied to the model. The interaction detection results of the three models are provided in Table III.

For the more-affected hands, the macro average MCCs when using only the Home dataset and both datasets were $0.34 \pm 0.19$ and $0.35 \pm 0.20$, respectively. The macro average F1-scores of the two conditions were both $0.73 \pm 0.17$. As for the micro average results, the MCCs in the former condition was 0.45 and the latter one was 0.46. The micro average F1-scores for the conditions were both 0.76. The results of the two conditions were very close for the more-affected hands.

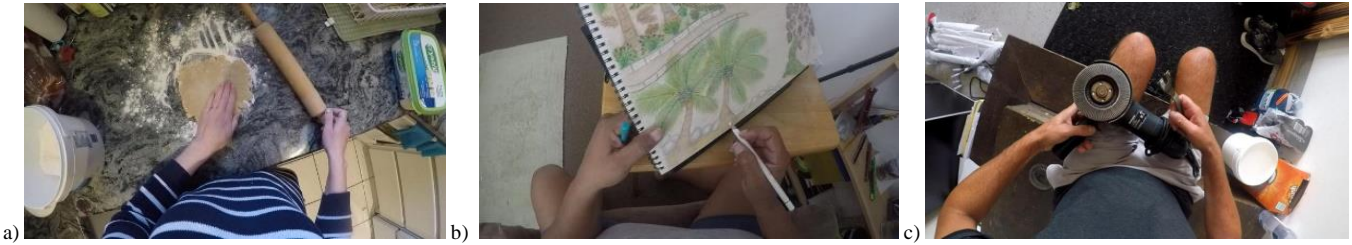

Fig. 2. Three examples from the Home dataset included (a) rolling dough, (b) drawing using a pencil, and (c) using tools.

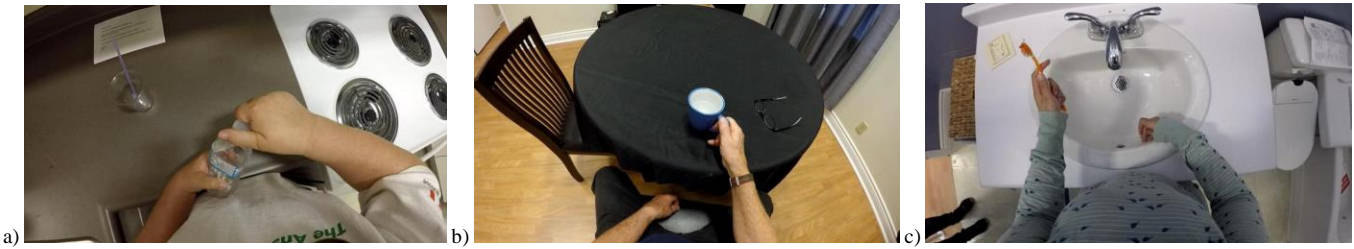

Figure 3. Three examples from the HomeLab dataset included (a) opening a bottle of water, (b) drinking from a mug, and (c) brushing teeth.

TABLE III
MACRO AND MICRO AVERAGE RESULTS OF LEAVE-ONE-SUBJECT-OUT-CROSS-VALIDATION (LOSOCV) FOR THE HAND-OBJECT INTERACTION DETECTION USING THE RANDOM FOREST CLASSIFIER, THE SLOWFAST NETWORK, AND THE HAND OBJECT DETECTOR IN THE TWO CONDITIONS: HOME DATASET AND BOTH DATASETS.

| Model | Average Type | Condition | More-Affected Hand | | | | | Less-Affected Hand | | | | | Overall (Both Hands) | | | | |
|---|---|---|---|---|---|---|---|---|---|---|---|---|---|---|---|---|---|
| | | | M | F | P | R | A | M | F | P | R | A | M | F | P | R | A |
| Random forest classifier | Macro average | Home | 0.34 ± 0.19 | 0.73 ± 0.17 | 0.68 ± 0.19 | 0.86 ± 0.17 | 0.72 ± 0.13 | 0.38 ± 0.18 | 0.78 ± 0.14 | 0.83 ± 0.10 | 0.77 ± 0.20 | 0.74 ± 0.11 | 0.37 ± 0.20 | 0.78 ± 0.11 | 0.76 ± 0.13 | 0.81 ± 0.13 | 0.73 ± 0.09 |
| | | Both datasets | 0.35 ± 0.20 | 0.73 ± 0.17 | 0.69 ± 0.19 | 0.85 ± 0.17 | 0.73 ± 0.13 | 0.39 ± 0.18 | 0.78 ± 0.14 | 0.84 ± 0.10 | 0.76 ± 0.20 | 0.74 ± 0.11 | 0.38 ± 0.20 | 0.78 ± 0.11 | 0.77 ± 0.13 | 0.81 ± 0.14 | 0.73 ± 0.09 |
| | Micro average | Home | 0.45 | 0.76 | 0.69 | 0.85 | 0.72 | 0.42 | 0.84 | 0.85 | 0.83 | 0.77 | 0.44 | 0.81 | 0.78 | 0.84 | 0.75 |
| | | Both datasets | 0.46 | 0.76 | 0.70 | 0.84 | 0.73 | 0.42 | 0.84 | 0.86 | 0.82 | 0.77 | 0.45 | 0.81 | 0.78 | 0.83 | 0.75 |
| SlowFast | Macro average | Home | 0.30 ± 0.26 | 0.71 ± 0.21 | 0.67 ± 0.24 | 0.81 ± 0.18 | 0.69 ± 0.13 | 0.32 ± 0.23 | 0.78 ± 0.14 | 0.79 ± 0.16 | 0.78 ± 0.16 | 0.72 ± 0.15 | 0.31 ± 0.21 | 0.75 ± 0.15 | 0.74 ± 0.17 | 0.79 ± 0.16 | 0.70 ± 0.12 |
| | | Both datasets | 0.32 ± 0.20 | 0.71 ± 0.19 | 0.68 ± 0.23 | 0.80 ± 0.18 | 0.69 ± 0.12 | 0.31 ± 0.18 | 0.79 ± 0.13 | 0.80 ± 0.14 | 0.81 ± 0.17 | 0.73 ± 0.10 | 0.31 ± 0.12 | 0.76 ± 0.12 | 0.74 ± 0.14 | 0.80 ± 0.15 | 0.71 ± 0.08 |
| | Micro average | Home | 0.39 | 0.74 | 0.67 | 0.83 | 0.69 | 0.33 | 0.80 | 0.84 | 0.76 | 0.72 | 0.36 | 0.77 | 0.76 | 0.79 | 0.71 |
| | | Both datasets | 0.38 | 0.73 | 0.67 | 0.81 | 0.69 | 0.32 | 0.80 | 0.84 | 0.77 | 0.72 | 0.36 | 0.77 | 0.76 | 0.79 | 0.71 |
| Hand Object Detector[†] | Macro average | | 0.50 ± 0.23 | 0.76 ± 0.18 | 0.75 ± 0.21 | 0.83 ± 0.17 | 0.76 ± 0.14 | 0.58 ± 0.18 | 0.87 ± 0.08 | 0.89 ± 0.09 | 0.86 ± 0.11 | 0.83 ± 0.09 | 0.54 ± 0.19 | 0.83 ± 0.08 | 0.83 ± 0.11 | 0.85 ± 0.11 | 0.80 ± 0.08 |
| | Micro average | | 0.49 | 0.77 | 0.72 | 0.83 | 0.74 | 0.60 | 0.90 | 0.90 | 0.90 | 0.85 | 0.55 | 0.84 | 0.82 | 0.87 | 0.80 |

M: Matthews correlation coefficient (MCC), F: F1-score, P: Precision, R: Recall, and A: Accuracy.
[†] The Hand Object Detector here was directly applied to the testing set without any retraining.

For the less-affected hands, the macro average MCC in the conditions using only the Home dataset was 0.38 ± 0.18 and the one using both datasets was 0.39 ± 0.18. The macro average F1-scores of the two conditions were both 0.78 ± 0.14. The micro average results in the two conditions were both 0.42 for MCCs and 0.84 for F1-scores.

For the combined results including data from each hand (overall results), the macro and micro average MCCs and F1-scores were also similar in the two conditions. The macro average MCCs for using the Home dataset and both datasets were 0.37 ± 0.20 and 0.38 ± 0.20, respectively. The micro average MCC for the former condition was 0.44 and for the latter one was 0.45. As for the average F1-scores in the two conditions, the macro ones were both 0.78 ± 0.11 and the micro ones were both 0.81. Using both datasets in the random forest classifier had slightly higher MCCs for either hand than using only the Home dataset.

*2. SlowFast Network*

The average number of testing instances in the two conditions was 462 ± 403 instances (66% ± 16% interaction and 34% ± 16% no interaction). The average numbers of training instances using the Home and both datasets were 7,627 ± 347 instances (65% ± 2% interaction and 35% ± 1% no interaction) and 13,665 ± 605 instances (61% ± 1% interaction and 39% ± 1% no interaction), respectively. The average number of validation instances was 1,619 ± 73 instances (58% ± 2% interaction and 42% ± 2% no interaction) for the former condition and 1,505 ± 75 instances (55% ± 2% interaction and 45% ± 2% no interaction) for the latter one.

In the macro average results for the more-affected hands using the Home dataset and both datasets, the MCCs were 0.30 ± 0.26 for the former and 0.32 ± 0.20 for the latter, and the F1-scores were 0.71 ± 0.21 and 0.71 ± 0.19, respectively. For the micro average results, the MCC and F1-score were 0.39 and 0.74 using only the Home dataset, and 0.38 and 0.73 using both datasets. Using both datasets had higher macro average MCCs to detect interactions of the more-affected hands.

For the less-affected hands, the macro average results for using one and two datasets had MCC 0.32 ± 0.23 for the former and 0.31 ± 0.18 for the latter, and F1-scores were 0.78 ± 0.14 and 0.79 ± 0.13, respectively. The macro average results were similar in the two conditions. The micro average MCC was 0.32 for using only the Home dataset and was 0.33 for using both datasets. The micro average F1-scores were 0.80 in both conditions.

In the overall results, the macro average results were also similar in the two conditions. The macro average MCCs for using the one and two datasets were 0.31 ± 0.21 and 0.31 ± 0.12, respectively. The macro average F1-scores were 0.75 ± 0.15 for the former and 0.76 ± 0.12 for the latter. The micro average MCCs were both 0.36 and the micro average F1-sccores were 0.77 in the two conditions.

TABLE IV
MACRO AND MICRO AVERAGE RESULTS OF LEAVE-ONE-SUBJECT-OUT-CROSS-VALIDATION (LOSOCV) FOR THE HAND ROLE CLASSIFICATION USING THE RANDOM FOREST CLASSIFIER, THE SLOWFAST NETWORK, AND THE HAND OBJECT DETECTOR IN THE TWO CONDITIONS: HOME DATASET AND BOTH DATASETS.

| Model | Average Type | Condition | More-Affected Hand | | | | | Less-Affected Hand | | | | | Overall (Both Hands) | | | | |
|---|---|---|---|---|---|---|---|---|---|---|---|---|---|---|---|---|---|
| | | | M | F | P | R | A | M | F | P | R | A | M | F | P | R | A |
| Random forest classifier | Macro average | Home | 0.03 ± 0.07 | 0.88 ± 0.20 | 0.83 ± 0.23 | 0.99 ± 0.03 | 0.82 ± 0.22 | 0.02 ± 0.05 | 0.82 ± 0.18 | 0.73 ± 0.23 | 0.99 ± 0.01 | 0.73 ± 0.23 | 0.04 ± 0.06 | 0.86 ± 0.08 | 0.78 ± 0.14 | 0.99 ± 0.01 | 0.77 ± 0.13 |
| | | Both datasets | 0.06 ± 0.13 | 0.86 ± 0.20 | 0.84 ± 0.22 | 0.93 ± 0.09 | 0.79 ± 0.22 | 0.08 ± 0.10 | 0.81 ± 0.17 | 0.74 ± 0.23 | 0.97 ± 0.03 | 0.73 ± 0.21 | 0.08 ± 0.11 | 0.85 ± 0.08 | 0.78 ± 0.14 | 0.94 ± 0.01 | 0.75 ± 0.12 |
| | Micro average | Home | -0.03 | 0.91 | 0.85 | 0.97 | 0.83 | 0.05 | 0.86 | 0.75 | 0.99 | 0.75 | -0.01 | 0.88 | 0.80 | 0.98 | 0.79 |
| | | Both datasets | -0.04 | 0.85 | 0.85 | 0.86 | 0.75 | 0.13 | 0.85 | 0.76 | 0.97 | 0.75 | 0.01 | 0.85 | 0.80 | 0.92 | 0.75 |
| Random forest classifier with Weighted Loss | Macro average | Both datasets | 0.12 ± 0.18 | 0.86 ± 0.20 | 0.84 ± 0.23 | 0.93 ± 0.07 | 0.79 ± 0.22 | 0.11 ± 0.13 | 0.82 ± 0.17 | 0.74 ± 0.23 | 0.96 ± 0.01 | 0.74 ± 0.21 | 0.12 ± 0.13 | 0.85 ± 0.08 | 0.79 ± 0.14 | 0.94 ± 0.05 | 0.76 ± 0.12 |
| | Micro average | | 0.02 | 0.88 | 0.86 | 0.9 | 0.79 | 0.21 | 0.86 | 0.77 | 0.97 | 0.76 | 0.11 | 0.87 | 0.81 | 0.93 | 0.77 |
| SlowFast | Macro average | Home | 0.00 ± 0.08 | 0.85 ± 0.20 | 0.83 ± 0.23 | 0.93 ± 0.12 | 0.78 ± 0.23 | 0.10 ± 0.27 | 0.82 ± 0.16 | 0.75 ± 0.22 | 0.96 ± 0.06 | 0.73 ± 0.21 | 0.06 ± 0.25 | 0.85 ± 0.10 | 0.78 ± 0.15 | 0.94 ± 0.07 | 0.75 ± 0.15 |
| | | Both datasets | 0.03 ± 0.15 | 0.84 ± 0.20 | 0.84 ± 0.22 | 0.91 ± 0.12 | 0.78 ± 0.18 | 0.02 ± 0.13 | 0.81 ± 0.17 | 0.74 ± 0.22 | 0.94 ± 0.06 | 0.72 ± 0.21 | 0.04 ± 0.14 | 0.84 ± 0.08 | 0.79 ± 0.13 | 0.92 ± 0.07 | 0.74 ± 0.12 |
| | Micro average | Home | -0.07 | 0.89 | 0.86 | 0.91 | 0.8 | 0.19 | 0.86 | 0.77 | 0.96 | 0.76 | 0.08 | 0.87 | 0.81 | 0.94 | 0.78 |
| | | Both datasets | 0.00 | 0.87 | 0.87 | 0.88 | 0.78 | 0.04 | 0.83 | 0.76 | 0.93 | 0.72 | 0.01 | 0.85 | 0.81 | 0.90 | 0.75 |
| Hand Object Detector | Macro average | Home | 0.00 ± 0.04 | 0.88 ± 0.21 | 0.83 ± 0.23 | 0.99 ± 0.02 | 0.82 ± 0.23 | 0.00 ± 0.07 | 0.83 ± 0.15 | 0.74 ± 0.21 | 0.99 ± 0.03 | 0.74 ± 0.20 | 0.00 ± 0.06 | 0.86 ± 0.09 | 0.78 ± 0.14 | 0.98 ± 0.02 | 0.77 ± 0.14 |
| | | Both datasets | -0.02 ± 0.10 | 0.78 ± 0.20 | 0.83 ± 0.23 | 0.80 ± 0.14 | 0.68 ± 0.19 | -0.06 ± 0.13 | 0.71 ± 0.18 | 0.72 ± 0.22 | 0.73 ± 0.19 | 0.62 ± 0.17 | -0.04 ± 0.09 | 0.75 ± 0.12 | 0.77 ± 0.14 | 0.76 ± 0.15 | 0.65 ± 0.13 |
| | Micro average | Home | -0.01 | 0.92 | 0.86 | 0.99 | 0.85 | 0.02 | 0.85 | 0.75 | 0.99 | 0.74 | 0.01 | 0.88 | 0.80 | 0.99 | 0.79 |
| | | Both datasets | -0.07 | 0.80 | 0.85 | 0.76 | 0.67 | -0.05 | 0.70 | 0.73 | 0.66 | 0.57 | -0.04 | 0.75 | 0.79 | 0.71 | 0.61 |

M: Matthews correlation coefficient (MCC), F: F1-score, P: Precision, R: Recall, and A: Accuracy.

### 3. Hand Object Detector

The macro average MCCs were 0.50 ± 0.23 for the more-affected hands and 0.58 ± 0.18 for the less-affected hands, and macros average F1-scores were 0.76 ± 0.18 for the former and 0.87 ± 0.08 for the latter. As for the overall results, the macro average MCC and F1-score were 0.54 ± 0.19 and 0.83 ± 0.08, respectively. In the macro average results between the three models, the Hand Object Detector had the highest MCCs and F1-scores for either hand. In the micro average results, the MCCs and F1-scores were 0.49 and 0.77 for the more-affected hands, 0.60 and 0.90 for the less-affected hand, and 0.55 and 0.84 for the combined results.

### 4. Statistical Results

In the hand-object interaction detection, when using the random forest classifier for either hand and the SlowFast network for the more-affected hands, including the HomeLab dataset during training led to a higher macro average MCC than using the Home dataset alone. Therefore, the results using both datasets were used to compute the between-group differences of the interaction detection using the three models. Based on the results from a Shapiro–Wilk test, the F1-scores of either hand were not normally distributed (p-value<0.05) and MCCs were normally distributed. The Friedman test was applied to the F1-scores and the repeated measures ANOVA was used for the MCCs.

Significant between-group differences of MCCs were found in the more-affected hands (p<0.01), less-affected hands (p<0.01), and overall results (p<0.01). Tukey's test was carried out as the post hoc test and the Hand Object Detector had significantly higher MCCs than the SlowFast network for the more-affected hands, and also higher than both other models for the less-affected hands and combined results.

Significant between-group differences of the F1-scores were found for the more-affected hands (p<0.05), less-affected hands (p<0.01), and overall results (p<0.01). Subsequently, the Wilcoxon signed rank test was carried out as a post hoc test and significance differences were found between the Hand Object Detector and the SlowFast network for the more-affected hands, and between the Hand Object Detector and both other models for the less-affected hands and combined results.

*C. Hand Role Results*

*1. Random Forest Classifier*

The testing set in the two conditions was the same and the average number of instances was 3,645 ± 3,153 instances (23% ± 14 % manipulation and 77% ± 14% stabilization). The average numbers of training instances using the Home and both datasets were 72,909 ± 3,153 instances (20% ± 1 % manipulation and 80% ± 1% stabilization) and 106,690 ± 3,694 instances (30% ± 1 % manipulation and 70% ± 1% stabilization), respectively.

For the more-affected hands, the macro average MCCs and F1-scores were 0.03 ± 0.07 and 0.88 ± 0.20 for using the Home dataset and 0.06 ± 0.13 and 0.86 ± 0.20 for using both datasets (Table IV). The micro average MCCs and F1-scores were -0.03 and 0.91 for the former and -0.04 and 0.85 for the latter. Using both datasets had higher macro average MCC and this suggested that including the HomeLab dataset could classify more-affected hand roles of unseen stroke survivors better than excluding the dataset. However, the micro average MCCs were both below zero in both conditions, which meant that the random forest classifier did not classify the more-affected hand roles well in general.

As for the less-affected hands, the macro average MCCs and F1-scores were 0.02 ± 0.05 and 0.82 ± 0.18 for using the Home dataset and 0.08 ± 0.10 and 0.81 ± 0.17 for using both datasets. The micro average MCCs and F1-scores were 0.05 and 0.86 for the former condition and 0.13 and 0.85 for the latter one. The macro and micro average MCCs supported that including the HomeLab dataset could classify the hand roles of less-affected hands better than using the Home dataset alone.

In the combined results for each hand (overall results), using both datasets had higher macro and micro average MCCs, which were 0.08 ± 0.11 and 0.01, compared to using only the Home dataset, which were 0.04 ± 0.06 and -0.01. The macro and micro average F1-scores were 0.86 ± 0.08 and 0.88 for using the Home dataset and 0.85 ± 0.08 and 0.85 for using both datasets.

In the hand role classification, the highest macro average MCC for the overall results was observed when using the random forest classifier and utilizing both datasets. Therefore, this condition was selected for an additional investigation. A modification was carried out to improve the performance by applying a weighted loss with a ratio of 20. By doing so, manipulation, which had a smaller number of instances, could get more attention during training. The macro and micro average MCC reached 0.12 ± 0.18 and 0.02 for the more-affected hands, 0.11 ± 0.13 and 0.21 for the less-affected hands, and 0.12 ± 0.13 and 0.11 for the combined results, which were all higher after applying the weighted loss (Table IV). This showed that applying a weighted loss to classify hand roles of stroke survivors was beneficial.

*2. SlowFast Network*

The average number of instances in the testing set in the two conditions was 603 ± 445 instances (19% ± 13% manipulation and 81% ± 13% stabilization). The average numbers of training instances using the Home and both datasets were 10,367 ± 451 instances (17% ± 1% manipulation and 83% ± 1% stabilization) and 16,586 ± 604 instances (26% ± 1% manipulation and 74% ± 1% stabilization), respectively. The average number of validation instances was 1,690 ± 103 instances (14% ± 1% manipulation and 86% ± 1% stabilization) for both conditions.

For the more-affected hands, the macro average MCCs and F1-scores were 0.00 ± 0.08 and 0.85 ± 0.20 for using the Home dataset and 0.03 ± 0.15 and 0.84 ± 0.20 for using both datasets. The micro average MCCs and F1-scores were -0.07 and 0.89 for the former condition and 0.00 and 0.87 for the latter condition. Including the HomeLab dataset had higher macro and micro average MCCs than using only the Home dataset.

For the less-affected hands, the macro average MCCs and F1-scores of using the Home dataset alone, which were 0.10 ± 0.27 and 0.82 ± 0.16, were higher than using both datasets, which were 0.02 ± 0.13 and 0.81 ± 0.17. The micro average MCC and F1-score were also higher when using the Home dataset alone, which were 0.19 and 0.86, than using both datasets, which were 0.04 and 0.83.

In the overall results, the macro average MCCs and F1-scores were higher when using the Home dataset, which were 0.06 ± 0.25 and 0.85 ± 0.10, than using both datasets, which were 0.04 ± 0.14 and 0.84 ± 0.08. The micro average MCC and F1-score of using the Home dataset were 0.08 and 0.87 and both were also higher than using both datasets, which were 0.01 and 0.85.

*3. Hand Object Detector*

The average number of testing instances was 4,156 ± 3,351 instances (22% ± 14% manipulation and 78% ± 14% stabilization). The average number of training instances were 72,908 ± 3,532 instances for using the Home dataset only (20% ± 1% manipulation and 80% ± 1% stabilization) and 115,331 ± 4,369 instances (31% ± 1% manipulation and 69% ± 1% stabilization) for using both datasets. The average numbers of validation instances were 10,223 ± 551 instances (19% ± 2% manipulation and 81% ± 2% stabilization) for the two conditions.

The macro average MCCs and F1-scores for more-affected hands were 0.00 ± 0.04 and 0.88 ± 0.21 for using the Home dataset and -0.02 ± 0.10 and 0.78 ± 0.20 for using both datasets. The micro average MCCs of the two conditions were both below zero and the micro average F1-scores were 0.92 for the former and 0.80 for the latter.

For the less-affected hands, the macro average MCCs and F1-scores were 0.00 ± 0.07 and 0.83 ± 0.15 for using the Home dataset and -0.06 ± 0.13 and 0.71 ± 0.18 for using both datasets. The micro average MCC and F1-score were 0.02 and 0.85 for the former and -0.05 and 0.70 for the latter.

The overall results were similar to the results of either hand that using Home dataset alone had higher macro average MCCs and F1-scores, which were 0.00 ± 0.06 and 0.86 ± 0.09, compared with using both datasets, which were -0.04 ± 0.09 and 0.75 ± 0.12. The micro average MCCs and F1-scores were 0.01 and 0.88 for the former and -0.04 and 0.75 for the latter.

*4. Statistical Results*

In the hand role classification, the MCCs and F1-scores of each participant without applying a weighted loss were compared between the three models. The results of either hand in the two conditions were not as consistent as the interaction detection. Here, the condition that used both datasets was again selected to investigate the between-group differences since the highest macro average MCC happened in this condition using the random forest classifier. The results of Shapiro–Wilk test showed that the MCCs and F1-scores of the more-affected and less-affected hands were not normally distributed (p-value<0.05) and the measures of the overall results were normally distributed. The Friedman test was applied to the more-affected and less-affected hands and the repeated measures ANOVA was used for the overall results.

Significant between-group differences in MCCs were found in the less-affected hands (p<0.01) and overall results (p<0.01). For the less-affected hands, the Wilcoxon signed-rank test was used as the post hoc test and the random forest classifier had significant higher MCCs than the other two models. As for the overall results (each hand), the Tukey's test was carried out as the post hoc test and the random forest classifier had significantly higher MCCs than the Hand Object Detector.

Significant between-group differences of F1-scores were found in the overall results (p<0.001). The random forest classifier and the SlowFast network were both found significantly higher F1-scores compared with the Hand Object Detector using the Tukey's test as the post hoc test. The F1-scores between the random forest classifier and the SlowFast network had no significant difference.

## IV. Discussion

This study used automated analysis of egocentric videos to analyze for the first time the hand function of stroke survivors in their home environment. Detecting hand-object interactions was demonstrated to be feasible, corroborating earlier findings in a laboratory environment [25]. The Hand Object Detector had the highest macro and micro average MCCs and F1-scores for the hand-object interaction detection. The detection of hand roles was found to be more challenging using analogous methods, and will warrant additional investigations with other approaches.

Inspection of the interaction detection results suggests that a closed hand posture was generally predicted as an interaction, which might generate false predictions among stroke survivors with severe to moderate hand function impairment, for whom closed fists are common even during resting. In the future, including more training instances from post-stroke individuals with closed hand shapes may improve the detection of hand-object interaction after stroke. As for hand roles, they were hard to classify due to the difficulty of distinguishing finger movements, which were the major difference between manipulation and stabilization. A manipulation includes more finger movements than a stabilization. Object movement or background changes might be a result of arm movements rather than finger movements, such as when an object was held during walking. Applying a weighted loss to the random forest classifier was beneficial for the hand role classification. Pose estimation methods [60-62] may provide a greater ability to tracks finger movements and could be a worthwhile avenue of investigation to better distinguish manipulation from stabilization.

In the comparison of the two conditions, the macro average results were very close and slightly higher when using both datasets than when using the Home dataset alone for the hand-object interaction detection. The results suggest that including a dataset that has a variety of examples of manipulating a set of objects in a controlled laboratory setting is beneficial to improving performance in an uncontrolled environment. In a laboratory setting, video quality and tasks carried out can be controlled to achieve adequate illumination and a clear view of the hands. These factors can ensure that videos are usable to train an interaction detection classifier. However, whether the improved performance of interaction detection comes from a larger number of training instances, a greater variety of controlled tasks, better lighting, or less camera motion during the short tasks, which are factors when including the HomeLab dataset, is out of the study scope and is left as a future question.

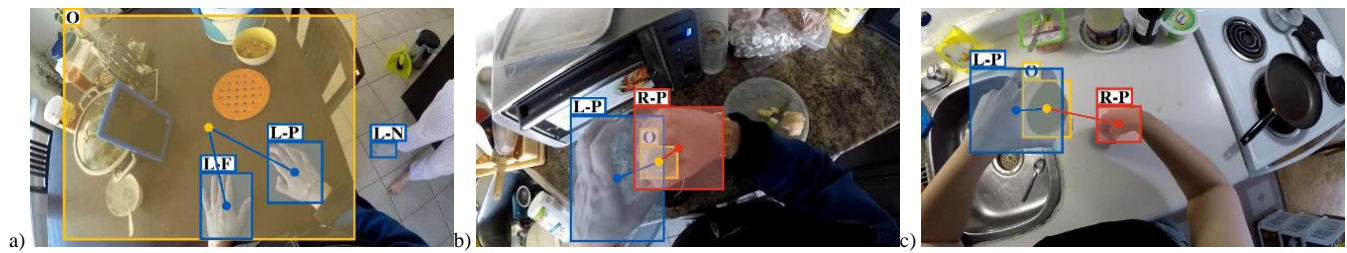

Figure 4. Examples of false hand-object detections using the Hand Object Detector: (a) false prediction on right hand and feet (b) left hand is close to an object but not in contact (c) right hand has a closed hand shape but is not in contact with any object.

In the results of the hand-object interaction for the more-affected, less-affected hands, and combined results, the macro average MCCs and F1-scores of the Hand Object Detector were significantly higher than those of the SlowFast network, which could be caused by the difference in the total number of training instances. The Hand Object Detector operates at the frame level, whereas SlowFast requires sequences as inputs, leading to a lower number of training instances. The Hand Object Detector was trained on more than 100,000 frames with hand contacts from healthy individuals, which is larger than our datasets and also demonstrates that using a dataset from healthy individuals can be beneficial to detecting the hand-object interactions of an unseen stroke survivor. Nonetheless, the Hand Object Detector predicted the interactions of the less-affected hands better, which had higher hand function that is closer to uninjured individuals. In the future, data augmentation could be applied to better identify the interactions of more-affected hands. In this study, factors that may have contributed to the different results between classifiers include the use of deep learning versus the manually selected features, the fact that some models used transfer learning whereas others were trained from scratch, and the different types of information leveraged by the three models. The Hand-Object Detector used information about objects near hands to detect hand contact and performed the best in the study. The SlowFast network was retrained from an activity recognition model, however, its performance was worse than the random forest classifier, which was trained from scratch. A possible explanation for this finding is that the selected features in the random forest classifier included additional information compared to SlowFast, such as differentiating hand and non-hand regions and using background motion. One contribution of this study is to provide insight into the factors that are beneficial for detecting the hand-object interactions of stroke survivors.

A majority of false interaction detections using the Hand Object Detector happened due to a false detection of hand side or hand location or when a hand was close to an object (Figure 4). These two types of failures were also reported in [55]. The participants who kept their fingers flexed in a closed hand shape most of the time during ADLs had higher false positive rates when detecting hand-object interactions, which meant that a closed hand shape was mostly predicted as an interaction (Figure 4c), even without contact with an object. In addition to these factors, the Hand Object Detector cannot differentiate the hands of the camera user from other people's hands, which also led to some false predictions. The Hand Object Detector is trained to detect contact, not interactions. Although it outperformed the other two models since hand contact covers most cases of hand-object interactions, some exceptions were found. For example, a hand resting on a table is not an interaction as defined in this study, however, there were instances of a hand in contact with the table being predicted as an interaction (Figure 4a).

In the hand role classification, the random forest classifier had the highest macro and micro average MCCs and F1-scores. One possible explanation for that was that the selected features compared the difference between hand and non-hand regions and included hand size changes over ten frames, which might capture finger movements better compared to the other two models. A stabilizer may have motion captured in the other two models that was not caused by finger movements. For example, in the case of a hand (stabilizer) that holds a sponge in place to clean a sink, the large movements for cleaning are caused by arm rather than hand or fingers. The need to identify whether the detected motion belongs to finger movements makes the hand role classification difficult. Applying pose estimation, which captures finger positions, may be a key step for the hand role classification in the future.

## V. CONCLUSION

Using automated analysis of egocentric videos to detect the hand-object interactions of stroke survivors at home is feasible. This study therefore provides a novel tool to evaluate independent hand use at home after stroke. Performance of the classifiers could be further improved in the future by conducting more training specific to impaired hand postures, such as the closed hand shapes that are associated with spasticity after stroke. Automatically identifying the role of the more-affected hand in bimanual interactions was found to be a more challenging task. Our results here provide a baseline for this novel problem. Possible avenues to improve on these results will include a greater focus on recognizing finger movements, such as by using pose estimation methods.

APPENDIX

A list of daily tasks carried out in the home simulation lab.

**Dinning Room**
1. Pick up a mug with liquid weight 350-400g.
2. Pick up a single sheet of standard paper from the round table.
3. Pick up a pencil from the round table.
4. Write a word with the pencil on the paper.
5. Pick up a book on the table, flip a few pages, and close it.
6. Pick up a credit card.
7. Swap the credit card on the mock machine.
8. Pick up a mobile phone.
9. Dial "123" on the mobile phone.
10. Use a key to open a lock.
11. Pick up a coin from the table and put it into a piggy bank.
12. Pick up a dice and toss it.
13. Pick up a nut from the table and screw it into the matching screw.
14. Open and close a Ziploc bag filled with 5 Golf balls on the round table.
15. Open web browser on a tablet, and type "google.ca".
16. Open and pick up a full pop can (355 ml) at the round table.
17. Pour 0.5 L of water from a bottle to the coffee cup.
18. Pretend to drink from the coffee cup.
19. Pick up and eat potato chips from a container.
20. Cut banana and eat with fork (banana already peel).
21. Clean hands with tissue.
22. Put your hand on the table and look around for 20 seconds.
23. Put your hand on side and look around for 20 seconds.
24. Right hand quick waves in the dinning room for 20 seconds.
25. Left hand quick waves in the dinning room for 20 seconds.

**Kitchen**
26. Open a disposable water bottle and pour some water into a disposable cup.
27. Grab a straw and place it in the cup, then drink from the cup.
28. Unscrew the lid of a Jar on the kitchen counter.
29. Pick up the sponge at the sink.
30. Open and close a container.
31. Open the fridge.
32. Put hands by the sink and look around for 20 seconds.
33. Put hands on the side, stand by the sink area and look around for 20 seconds.
34. Put hands on the counter area and look around for 20 seconds.
35. Put hands on the side, stand by the counter and look around for 20 seconds.
36. Right hand quick waves in the kitchen for 20 seconds.
37. Left hand quick waves in the kitchen for 20 seconds.

**Washroom**
38. Wash hands with hand soap in the washroom.
39. Pick up the toothbrush and pretend to clean teeth in the washroom.
40. Open and close the pill organizer.
41. Replace the empty tissue roll in the washroom.
42. Put hands on the washroom sink and look around for 20 seconds.
43. Put hands on the side, stand by the washroom sink, and look around for 20 seconds.
44. Right hand quick waves in the washroom for 20 seconds.
45. Left hand quick waves in the washroom for 20 seconds.

**Living Room**
46. Open a folded newspaper, pretend to read and fold back the newspaper.
47. Turn on and off the TV using the remote.
48. Type "google.ca" using a keyboard.
49. Put hands on laps while sitting on the sofa and look around the 20 seconds.

50. Put hands on laps and cover the hands with a pillow while sitting on the sofa and look around the 20 seconds.
51. Right hand quick waves in the living room for 20 seconds.
52. Left hand quick waves in the living room for 20 seconds.

**Bedroom**
53. Open the door in the bedroom.
54. Pick up the T-shirt from the bed and hang it up in the bar.
55. Fold a towel and place it in the drawer.
56. Sit on the bed with hands on laps and look around for 20 seconds.
57. Sit on the bed with hands on laps and cover the hands with a pillow, and look around for 20 seconds.
58. Right hand quick waves in the bedroom for 20 seconds.
59. Left hand quick waves in the bedroom for 20 seconds.

**Hallway**
60. Open the door on the hallway.
61. Place oranges (tennis balls) on the bench into a plastic bag.
62. Put one hand behind the back and the other hand on the bench, look around the bench area for 20 seconds.
63. Put both hands on the bench, look around the bench area for 20 seconds.
64. Right hand quick waves for 20 seconds.
65. Left hand quick waves for 20 seconds.